\documentclass{article}

\usepackage{amsmath}
\usepackage{amssymb}
\usepackage{tabularx}
\usepackage{booktabs}
\usepackage{array}
\usepackage{algorithmic}
\usepackage[linesnumbered, vlined, ruled]{algorithm2e}
\usepackage{graphicx}

\usepackage{subfig}
\usepackage{caption}

\usepackage[final]{corl_2025} 

\title{IndoorBEV: Joint Detection and Footprint Completion of Objects via Mask-based Prediction in Indoor Scenarios for Bird's-Eye View Perception}

\author{
Haichuan Li$^{\dagger1}$ \quad
Changda Tian$^{\dagger2}$ \quad
Panos Trahanias$^{2}$ \quad
Tomi Westerlund$^{1}$
}

\begin{document}
\maketitle

\footnotetext[1]{Turku Intelligent Embedded and Robotics Systems lab, Faculty of Technology, University of Turku, Finland. \texttt{\{haicli, tovewe\}@utu.fi}} 
\footnotetext[2]{Institute of Computer Science, Foundation for Research and Technology–Hellas, Greece. \texttt{\{dada, trahania\}@ics.forth.gr}}
\footnotetext[3]{$\dagger$ Equal contribution.}

\begin{figure}[h!]
    \centering
    \includegraphics[width=\textwidth]{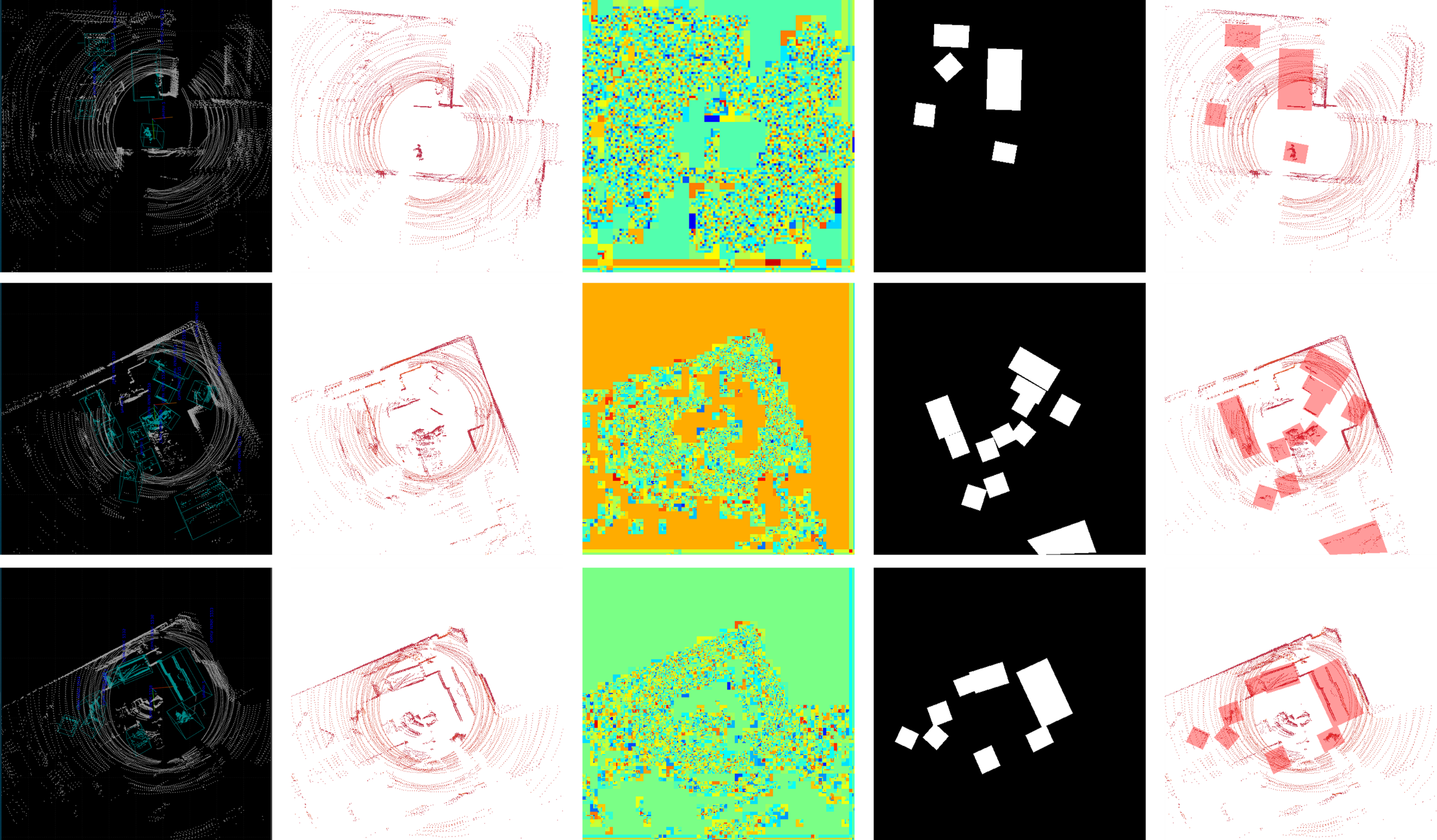}
    \caption{A unified high-performance IndoorBEV framework generates BEV representations and predicts object footprint masks directly from raw lidar point cloud data in indoor environments. The first column represents the raw 3D point clouds as each input. Next, the axis fusion results for each correlated input are presented. The third column shows the features maps generated by the backbone, followed by the predicted mask of each objects in each secnes in the next column. Finally, the last column shows the final BEV result of each input}
    \label{fig:teaser}
\end{figure}

\begin{abstract}
    Detecting diverse objects within complex indoor 3D point clouds presents significant challenges for robotic perception, particularly with varied object shapes, clutter, and the co-existence of static and dynamic elements where traditional bounding box methods falter. To address these limitations, we propose IndoorBEV, a novel mask-based Bird's-Eye View (BEV) method for indoor mobile robots. 
    In a BEV method, a 3D scene is projected into a 2D BEV grid which handles naturally occlusions and provides a consistent top-down view aiding to distinguish static obstacles from dynamic agents. The obtained 2D BEV results is directly usable to downstream robotic tasks like navigation, motion prediction, and planning. Our architecture utilizes an axis compact encoder and a window-based backbone to extract rich spatial features from this BEV map. A query-based decoder head then employs learned object queries to concurrently predict object classes and instance masks in the BEV space. This mask-centric formulation effectively captures the footprint of both static and dynamic objects regardless of their shape, offering a robust alternative to bounding box regression. We demonstrate the effectiveness of IndoorBEV on a custom indoor dataset featuring diverse object classes including static objects 
    and dynamic elements like robots and miscellaneous items, showcasing its potential for robust indoor scene understanding.

\end{abstract}



\section{Introduction}

The deployment of autonomous robots in indoor environments necessitates precise and real-time perception of their surroundings to ensure safe and efficient navigation. Lidar sensors have emerged as a pivotal technology in this domain, offering high-resolution 3D point cloud data that is robust to varying lighting conditions and capable of capturing intricate spatial details. However, transforming this unstructured point cloud data into actionable representations for tasks such as object detection, segmentation, and navigation remains a formidable challenge, particularly given the complexity and clutter often found indoors.

Recent advancements have sought to address these challenges. For instance,  MakeWay~\cite{xu2024makewayobjectawarecostmapsproactive} system introduces object-aware costmaps derived from lidar data to enhance proactive indoor navigation. Similarly, the LV-DOT framework~\cite{xu2025lvdotlidarvisualdynamicobstacle} leverages a fusion of lidar and visual data to improve dynamic obstacle detection and tracking in indoor settings. These approaches underscore the potential of integrating machine learning techniques with lidar data to enhance indoor perception.

Bird's-Eye View (BEV) representations naturally handles occlusions and provides a representation directly amenable to downstream robotic tasks like navigation and planning due to their ability to provide a top-down, spatially consistent view of the environment.~\cite{10.1007/978-3-031-20077-9_1,10321736}. However, adapting BEV methodologies to indoor environments is non-trivial. Indoor settings often present challenges such as limited sensor range, significant clutter from diverse object shapes, and frequent occlusions, which can degrade the performance of traditional BEV generation and perception techniques.

In this paper, we propose a novel environment perception framework, IndoorBEV. IndoorBEV provides real-time BEV perception tailored to indoor environments, focusing on joint object detection and footprint mask prediction directly from lidar point clouds. Our approach utilizes a dimension-fusion pre-processing. Initially, 3D point cloud data is processed by an encoder to fuse the axis information then to generate a 2D BEV grid. After that, this BEV grid is fed into a backbone which learns rich spatial context across the BEV grid. Crucially, instead of predicting traditional bounding boxes, our method employs a query-based decoder head which uses a set of learned object queries to interact with the BEV feature map. Each query is trained to predict the class, 3D pose attributes (position, yaw), and, most importantly, a pixel-wise instance mask representing the object's footprint within the BEV grid. This mask-centric formulation allows the algorithm to capture the precise 2D extent of diverse objects, offering a more flexible representation than bounding boxes for tasks like collision avoidance and path planning. The whole IndoorBEV algorithm is shown in ~\autoref{alg:IndoorBEV}.

\begin{figure}[ht]
    \centering
    \includegraphics[width=\textwidth]{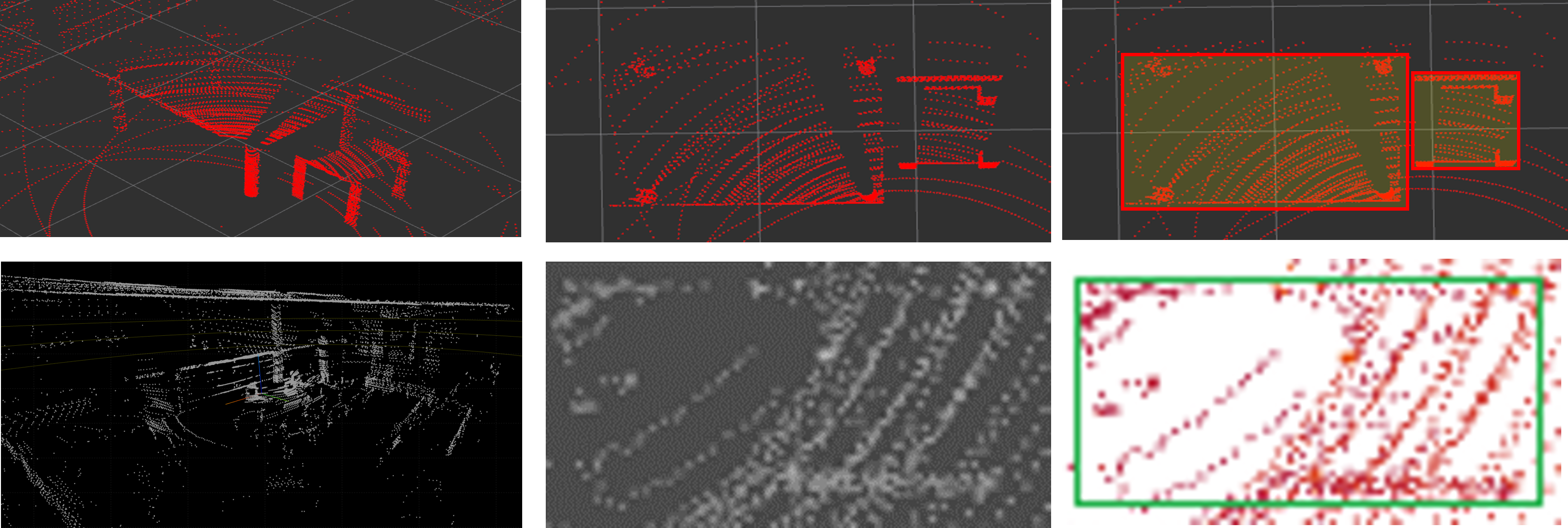}
    \caption{From left to right: Original point cloud. BEV point cloud without complete boundaries. Masked object with whole edges. The first row is in simulation, the second in the real world. }
    \label{fig:mask}
\end{figure}

Our IndoorBEV framework designed to operate in real-time, enabling seamless integration with robotic locomotion modules for dynamic navigation tasks for indoor environments. The primary contributions of this paper are as follows:
\begin{itemize}
    \item  We introduce a pioneering BEV-based indoor perception framework on mobile robots.
    \item A unified high-performance framework for generating BEV representations and predicting object footprint masks directly from raw lidar point cloud data in indoor environments.
    \item A query-based mask prediction head that avoids bounding box regression, enabling accurate footprint capture for diverse and complex indoor object shapes, see in Fig.~\ref{fig:mask}.
   \item A hybrid dataset combining simulated and real-world lidar data, facilitating effective training and evaluation of the proposed system.
\end{itemize}

\section{Related Work}

Bird's-Eye View representations have gained significant traction in 3D object detection, particularly in outdoor autonomous driving scenarios. These methods leverage the structured nature of lidar point clouds to project information into a BEV space, where convolutional neural networks can effectively reason about spatial relationships. One notable approach is BEV-MAE~\cite{lin2022bev}, which introduces a masked autoencoding framework for pre-training on lidar point clouds. By employing a BEV-guided masking strategy, BEV-MAE enhances the learning of robust representations for 3D object detection and achieves state-of-the-art performance on the nuScenes dataset.
Complementary to this, Mohapatra et al. propose BEVDetNet~\cite{mohapatra2021bevdetnet}, a real-time 3D object detection network specifically optimized for embedded systems. Their architecture processes lidar data directly in the BEV domain, demonstrating both computational efficiency and competitive accuracy, making it well-suited for practical deployment in autonomous vehicles. These methods exemplify the success of BEV-based frameworks in outdoor environments, where road scenes are relatively structured and object categories are consistent.

Traditional BEV detection methods often rely on regressing oriented bounding boxes~\cite{article,raut2023endtoend3dobjectdetection,10.1109/ITSC45102.2020.9294293}, which can be limited in capturing the fine-grained structure of objects with irregular shapes. To address this, Zhao et al. introduce MaskBEV~\cite{zhao2024maskbev}, a unified framework that reframes BEV object detection as a segmentation task. Instead of regressing bounding boxes, their method predicts a set of BEV instance masks that directly capture object footprints. This classification-based formulation facilitates the joint learning of object detection and map segmentation in a shared architecture. Although effective, MaskBEV has been evaluated solely on outdoor urban scenes.

In contrast, the use of BEV perception in indoor environments remains relatively underexplored. Indoor scenes present distinct challenges, including cluttered layouts, varied object geometries, and occlusions not typically encountered outdoors. An exception is the work by Wang et al. who propose BEV-SUSHI~\cite{wang2024bev}, a BEV-based multi-target multi-camera 3D detection and tracking framework designed for general indoor and outdoor surveillance applications. While BEV-SUSHI demonstrates promising results in multi-camera setups and emphasizes the generalizability of BEV perception across domains, its primary focus lies in tracking and multi-view aggregation rather than footprint prediction or joint detection and segmentation.

To the best of our knowledge, no prior work has directly addressed the perception of mobile robots in the BEV space specifically designed for indoor environments. This gap motivates our approach, which aims to leverage the advantages of BEV representations while overcoming the limitations of bounding box-based formulations in complex and cluttered indoor settings.



\begin{figure}[t]
    \centering
    \includegraphics[width=\textwidth]{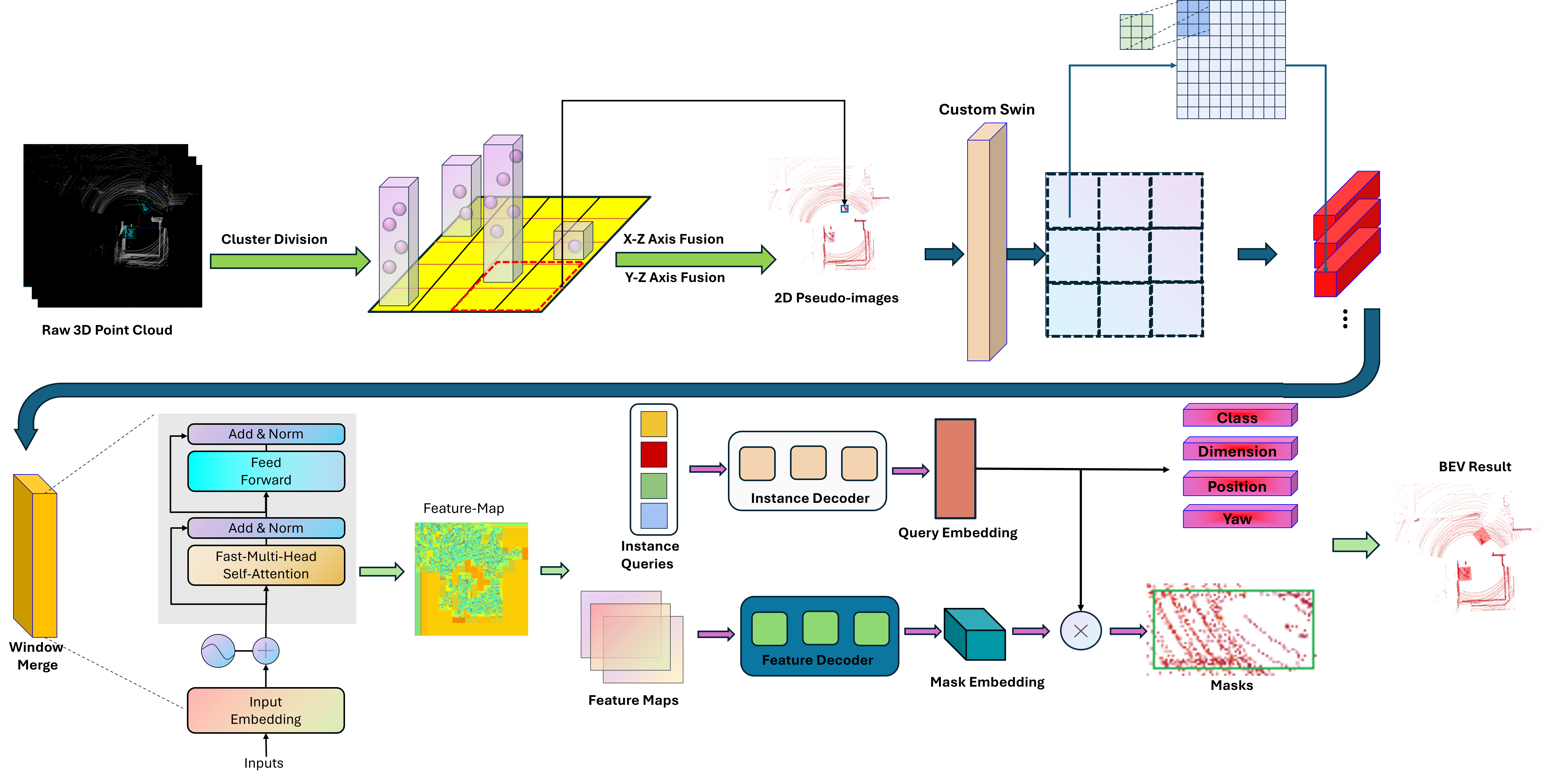}
    \caption{Overview of the \textit{IndoorBEV} framework. 
Lidar point clouds are divided into clusters then fused with the axis information features and projected to a BEV map.
The BEV features are processed by a window-based backbone to capture spatial context.
Learnable object queries interact with the BEV features through a mask-based decoder to predict object classes, positions, dimensions and dense BEV masks.}
    \label{fig:overview}
\end{figure}

\section{Method}
\label{sec:citations}

Our proposed framework, IndoorBEV, performs joint object detection and instance footprint mask prediction directly from raw lidar point clouds within a BEV representation. The predicted masks show the full edges of each instance, which is a significant advantage over the traditional 3D object detection methods. This method allows for object completion of arbitrary shapes, as the shape prior is provided by the ground truth annotation in the form of masks~\cite{chen2023pimaepointcloudimage}. The architecture follows three main components: an axis fusion encoder, a window-based backbone, and a mask-based decoder. An overview of the architecture is depicted in Figure.~\ref{fig:overview}

\subsection{Dimension Fusion Encoder}

The axis fusion encoder utilises our \textit{Axis-Fusion Point Cloud Compact Network (AF-PCCN)}. In AF-PCNN, the input 3D point cloud $\mathcal{P} = \{\mathbf{p}_i\}_{i=1}^{N}$ with $\mathbf{p}_i \in \mathbb{R}^d$ ($d = 3$ for the $X$, $Y$, $Z$ axis, or $d = 4$ if intensity is included), is first discretized into a numerous clusters based on predefined spatial ranges ($(x_\text{range}, y_\text{range}, z_\text{range})$) and clusters size ($v$). The divided clusters are mostly empty because of the sparsity of the point clouds, and the non-empty clusters in general have few points in them~\cite{Zhou2018}. In contrast of directly applying a PointNet~\cite{Qi2016PointNetDL}, which pads numerous zeros to a result to include each cluster the same number of points, we use a learnable parameters based network to fuse the $X-Z$ axis and $Y-Z$ axis features, and then compose the 2D results into a pseudo image. This ensures that empty clusters will be marked as noise. The resulting image is of size $(F \times H \times W)$, where $F$ is the embedding dimension of clusters, $H$ and $W$ are the number of clusters along the $Y$-axis and $X$-axis, respectively. These cluster features are then projected or scattered onto a 2D BEV grid of size by function
$\mathbf{F}_{\text{enc}} \in \mathbb{R}^{C_\text{enc} \times H \times W}$, where
$H = (y_\text{range}^{\max} - y_\text{range}^{\min})/{v},$ and $
W = (x_\text{range}^{\max} - x_\text{range}^{\min})/{v}.$
This forms an initial BEV feature map $(F_{bev} \in \mathbb{R}^{C_{enc} \times H \times W})$. Our encoder incorporates point coordinates and distance features during cluster feature computation.

\subsection{Window-based Backbone}
The initial BEV feature map ($F_{bev}$) is processed by a window-based backbone to capture long-range spatial dependencies and contextual information across the BEV plane. Specifically, we employ a CustomSwinTransformer architecture which is modified from SwinTransformer~\cite{hwang2022tutel} for BEV feature processing. This backbone consists of four stages, where each stage comprises Swin Transformer blocks applying windowed multi-head self-attention (W-MSA) and shifted-window multi-head self-attention (SW-MSA). To manage computational cost while retaining initial resolution, we configure the backbone with strides of $(1, 2, 2, 2)$ across the stages. This means the first stage operates at the input BEV resolution $(H \times W)$, while subsequent stages progressively downsample the spatial resolution by a factor of 2 and increase the feature channel dimension. We utilize a lightweight configuration with depths $(1, 1, 1, 1)$, number of attention heads $(2, 4, 8, 8)$ per stage, a window size of 7, and an MLP ratio of 2. Gradient checkpointing is enabled $(with\_cp=True)$ to reduce memory consumption during training. While the backbone can output feature maps from multiple stages $(out\_indices=(0, 1, 2, 3))$, only the feature map from the final stage 
$(F_{backbone} \in \mathbb{R}^{C_{backbone} \times H' \times W'})$, where ($H'$, $W'$) are the most downsampled dimensions. The final stage feature map is passed to the subsequent decoder head after upsampling.

\subsection{Query-Based Detection and Mask Head Decoder}

Inspired by query-based object detection frameworks like DETR~\cite{10.1007/978-3-030-58452-8_13} and Mask2Former~\cite{cheng2021mask2former}, we employ a mask-based decoder head (SimpleMaskHead) for simultaneous object classification, 3D attribute regression, and instance mask prediction. 

\textbf{Upsampling and Projection}: The final backbone feature map ($F_{backbone}$) is first upsampled back to the original BEV resolution ($H \times W$) and projected to a hidden dimension ($C_{hidden}$) using an adaptive upsampler module (AdaptiveUpsampler followed by $input\_proj$ and $input\_proj\_norm$ within the head). This results in a refined feature map 
$(F_{proj} \in \mathbb{R}^{C_{hidden} \times H \times W})$
which serves as the memory input for the decoder.

\textbf{Object Queries and mask Decoder}: A set of $(N_q)$ learnable object queries ($Q = {q_j}{j=1}^{N_q}$), represented as embeddings ($q_j \in \mathbb{R}^{C{hidden}}$) ($query\_embed$), are initialized. These queries are processed through a stack of standard transformer decoder layers. Each decoder layer performs self-attention among the queries and cross-attention between the queries and the projected BEV feature map ($F_{proj}$). This allows each query to aggregate information from the relevant spatial locations in the BEV map and specialize towards predicting a specific object instance. The output of the final decoder layer is a set of refined query embeddings ($Q' = {q'_j}{j=1}^{N_q}$).

\textbf{Prediction Heads}: Several prediction heads operate on the refined query embeddings (Q'):
\begin{itemize}
    \item \textit{Classification Head}: A linear layer predicts the class logits $(L_{cls} \in \mathbb{R}^{N_q \times N_{classes}})$ for each query, where $(N_{classes})$ includes foreground classes and a background/no-object class.
    \item \textit{Regression Heads}: Linear layers predict 3D attributes associated with each query: dimensions $(D \in \mathbb{R}^{N_q \times 3})$ (length, width, height), position $(P \in \mathbb{R}^{N_q \times 3})$ (x, y, z), and yaw $(Y \in \mathbb{R}^{N_q \times 1})$.
    \item \textit{Mask Prediction Head}: This head dynamically generates instance masks. First, an MLP ($mask\_embed\_head$) transforms each refined query embedding ($q'_j$) into a mask embedding ($m_j \in \mathbb{R}^{C{hidden}}$). The final mask logits ($M \in \mathbb{R}^{N_q \times H \times W}$) are computed via matrix multiplication between the mask embeddings and the projected BEV features: ($M_j = m_j \cdot F_{proj}$), where the multiplication occurs across the channel dimension and results are reshaped to the BEV grid dimensions. This allows each query to produce a unique spatial mask based on both its learned specialization and the input features.
\end{itemize}

\subsection{MuJoCo–Taichi-Based Lidar Simulation}
\label{sec:mujoco-lidar}

We develop a high-fidelity lidar simulator using MuJoCo~\cite{todorov2012mujoco} for environment modeling and Taichi~\cite{hu2019taichi} for massively parallel ray-casting.  
Procedural indoor scenes are generated with diverse object geometries, including planes, boxes, spheres, cylinders, capsules, and ellipsoids.

Given a sensor pose $\mathbf{T}_s \in \mathrm{SE}(3)$, rays are emitted using spherical coordinates $(\theta, \phi)$:
\begin{equation}
\mathbf{d}(\theta, \phi) = 
\begin{bmatrix}
\cos\phi \cos\theta \\
\cos\phi \sin\theta \\
\sin\phi
\end{bmatrix},
\quad
\mathbf{p}(t) = \mathbf{o} + t \mathbf{d},
\end{equation}
where $\mathbf{o}$ is the sensor origin and $t$ is the ray distance.

Each ray is transformed into object-local frames for efficient intersection tests. The intersection with each geometry reduces to solving analytic forms which is shown in Appendix ~\ref{app2}

The closest valid hit $t^\ast = \arg\min_{t>0} t$ for each ray is kept, and the hit points form the point cloud $\mathcal{P} = \{\mathbf{p}_i\}_{i=1}^N$ in the sensor frame.

Our Taichi implementation parallelizes ray tracing across all rays and objects, utilizing optimized memory preallocation, selective updates, and kernel fusion.  
The system maintains real-time performance even with dense lidar scans ($>$100k rays), supporting dynamic scene updates without reinitializing memory structures.
    
\section{Experiments}
This section details the simulation environment, datasets, training setup, implementation specifics, evaluation metrics, and presents the results of our proposed \emph{IndoorBEV} framework.

\subsection{Datasets}
\label{sec:datasets}

We evaluate our method on a custom hybrid dataset tailored for indoor robotic environments, comprising both simulated and real-world data.

\paragraph{Simulated Data.}  
Synthetic data is generated via our customized MuJoCo pipeline, producing diverse floor plans, object layouts (e.g., tables, chairs, shelves, or clutter), randomized object poses, and lidar noise models.  
For each frame, perfect ground-truth annotations are automatically extracted, including object classes, 3D poses, and dimension information (length, width, height) of the objects.  
The MuJoCo scenes are ray-traced with a high-performance Taichi-based lidar simulator to simulate realistic point cloud measurements.

\begin{figure}[ht]
    \centering
    \subfloat[Simulation Scene]{\includegraphics[height=3.5cm]{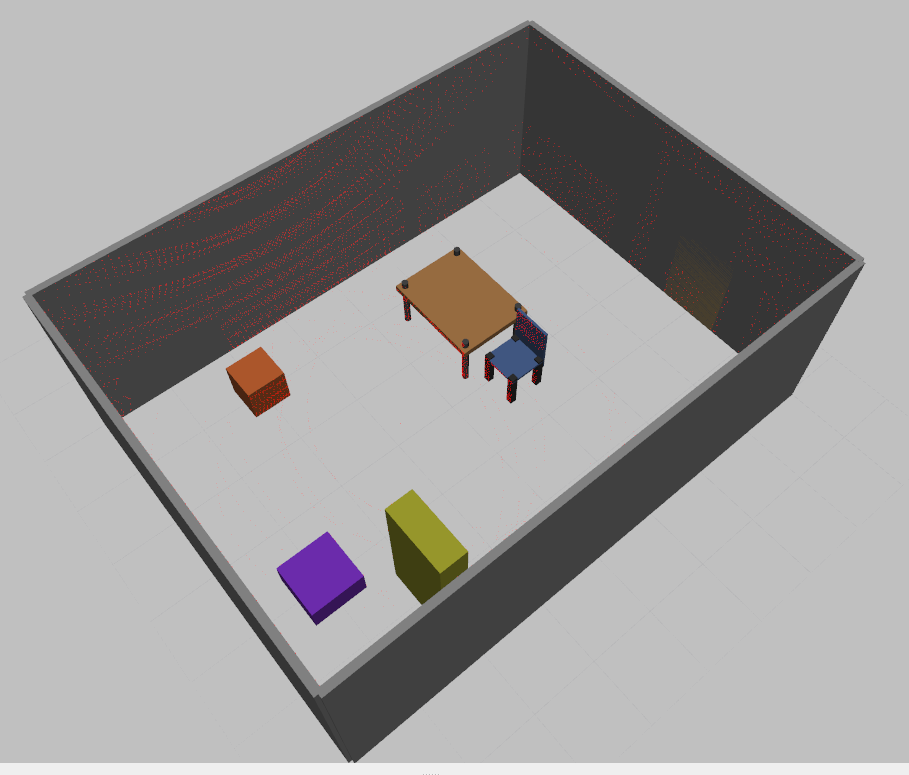}\label{subfig:sim}}
    \hfill
    \subfloat[Simulation Point Cloud]{\includegraphics[height=3.5cm]{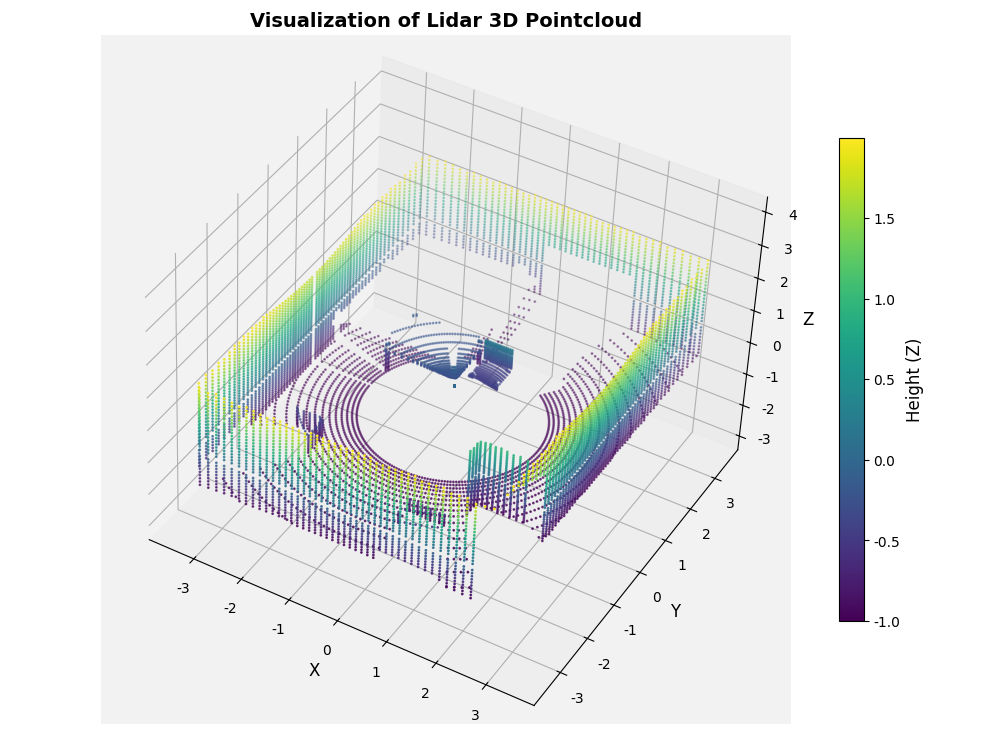}\label{subfig:point_cloud}}
    \hfill
    \subfloat[Data Collection Platform]{\includegraphics[height=3.5cm]{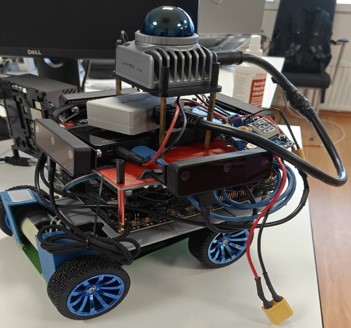}\label{subfig:indoor}}
    \caption{(a): Simulated indoor environment in MuJoCo. (b): 3D point cloud rendered from ray-traced lidar. (c): The mobile robot we used for the data collection}
    \label{fig:sim_results}
\end{figure}

\paragraph{Real-World Data.}  
Real-world scans are collected using a Livox MID-360 mounted on a mobile robot traversing offices, laboratories, and corridors show in Fig.~\ref{fig:sim_results}(c).  
Ground-truth annotations (object classes, 3D poses, and footprint masks) are generated using semi-automatic pipelines and manually verified via SUSTechPOINTS~\cite{9304562}.  
The dataset is split into approximately ${\sim}4.5\,\mathrm{k}$ frames for training, $1\,\mathrm{k}$ for validation, and $1\,\mathrm{k}$ for testing.

\paragraph{Data Format and Preprocessing.}  
Each frame is stored following a format inspired by the KITTI~\cite{Geiger2013} object detection benchmark:

\textit{Input Point Clouds:} Lidar scans are stored in \texttt{.bin} files, containing $N\times4$ NumPy arrays (x, y, z, intensity).
\textit{Labels:} Annotations follow KITTI’s \texttt{label\_2} format, recording the object type (e.g., Person, Table, or Chair), truncation, occlusion level, dimensions, 3D location, and yaw.
\textit{Calibration:} No sensor fusion calibration is required, as a single lidar sensor is used.

For BEV-based training, 3D labels are rasterized into 2D instance masks:
\begin{itemize}
    \item \textit{Rasterization:} Each object is projected to the BEV plane using its 3D dimensions, location, and yaw. The MaskRasterizer function in our project maps the object footprint to the BEV grid as a filled polygon assigned a unique instance ID.
    \item \textit{Label Formatting:} Masks and class labels are converted into padded tensors, matching the number of object queries, with auxiliary fields for 3D attributes (dimensions, position, yaw) to enable multi-task supervision.
\end{itemize}


This pipeline ensures diverse, challenging training data with both geometric and semantic variability, enhancing robustness for instance-aware BEV perception tasks.

\subsection{Simulation-based Data Generation and ROS2 Integration}
\label{sec:simulation}

To support large-scale training and domain adaptation, we develop a custom lidar simulation system using MuJoCo for physics-based 3D modeling and Taichi for high-performance ray tracing. The simulator procedurally generates indoor layouts with varied geometry, materials, and lighting. Each frame renders synthetic lidar scans by emitting rays over a dense spherical grid and computing intersections with scene objects. Moreover, we use a ROS2~\cite{ros2} system to record frames into bags.

Each beam direction $(\theta_i, \phi_i)$ yields a normalized vector $\mathbf{r}_i$, and the closest intersection point is obtained using shape-specific analytical solvers. Hit points are transformed to the lidar frame using inverse sensor poses and accumulated into a 3D point cloud.

\subsection{Training and Loss Function}
\label{sec:training}

We adopt a set-based training framework, following DETR~\cite{10.1007/978-3-030-58452-8_13}, using bipartite matching between $N_q$ predicted queries and ground-truth objects via the Hungarian algorithm~\cite{Kuhn1955Hungarian}.  
The matching cost considers classification, mask, and regression terms:
\begin{equation}
  \mathcal{C} =
    \alpha_{\text{cls}}(1-\hat{p}_{\text{cls}})
  + \alpha_{\text{dice}}\!\bigl(1-\text{Dice}(\hat{M},M)\bigr)
  + \alpha_{\text{foc}}\,\text{Focal}(\hat{M},M)
  + \alpha_{\text{reg}}\left\lVert\hat{\mathbf{b}}-\mathbf{b}\right\rVert_{1},
\end{equation}
where $\hat{p}_{\text{cls}}$ and $\hat{M}$ denote the predicted class probability and mask, respectively, and $\hat{\mathbf{b}}$ denotes 3D box parameters (dimensions, position, yaw).
For matched pairs, the final loss minimized is:
\begin{equation}
  \mathcal{L} =
      \lambda_{\text{cls}} \mathcal{L}_{\text{cls}}
    + \lambda_{\text{dice}}\mathcal{L}_{\text{dice}}
    + \lambda_{\text{mask}}\mathcal{L}_{\text{focal}}
    + \lambda_{\text{dim}}\lVert\hat{\mathbf{d}}-\mathbf{d}\rVert_{1}
    + \lambda_{\text{pos}}\lVert\hat{\mathbf{p}}-\mathbf{p}\rVert_{1}
    + \lambda_{\text{yaw}}\lVert\hat{\psi}-\psi\rVert_{1},
\end{equation}
where $\mathcal{L}_{\text{cls}}$ is the cross-entropy loss (with a down-weighted background class), $\mathcal{L}_{\text{dice}}$ and $\mathcal{L}_{\text{focal}}$ are mask losses, and regression losses apply L1 loss on dimensions, position, and yaw. Auxiliary losses are added at intermediate decoder layers. Weights are default set to $(2.0,5.0,2.0,0.1,0.1,0.1)$, with unmatched queries assigned a background weight of 0.1.

\subsection{Implementation Details}
\label{sec:implementation}

The cluster size is set to $v=0.02$\,m, with a perception workspace of $[-5,5]\times[-5,5]\times[-3,3]$\,m, resulting in a BEV grid of size $500\times500$.  
The encoder outputs features of dimension $C_\text{enc}=64$.  
The backbone is a lightweight CustomSwinTransformer, with four stages, strides $(1,2,2,2)$, and depths $(1,1,1,1)$.  
The mask-based decoder (SimpleMaskHead) uses $N_q=30$ object queries, hidden dimension $C_\text{hid}=64$, four transformer decoder layers, and 256 attention heads. The classification head predicts $N_\text{cls}=9$ classes.

Training is performed using the AdamW optimizer~\cite{loshchilov2018decoupled} with an initial learning rate of $5\times10^{-5}$, weight decay $10^{-4}$, and a cosine learning rate schedule with plateau warm-up.  
The algorithm were trained on five nodes each equips with two Intel Xeon processors, code name Cascade Lake, with 20 cores each running at 2,1\,GHz and contains four Nvidia A100 GPUs, for 12k iterations with a batch size of 16.  
Data augmentations include random flipping, rotation, range dropout, and global-scaling jitter.  
During inference, predictions are filtered using a class probability threshold of 0.8, and no non-maximum suppression is applied due to bipartite matching.

\subsection{Evaluation Metrics}
\label{sec:metrics}

We evaluate detection and segmentation in BEV space using standard metrics for object detection and instance segmentation, adapted for BEV mask predictions. For detection, we use Average Precision (AP) metrics calculated based on mask Intersection over Union (IoU) at thresholds of 0.25 and 0.50 (AP@0.25, AP@0.50). For segmentation, we use Mean Intersection over Union (mIoU) averaged over all foreground classes, comparing predicted masks to ground truth masks for correctly detected objects. For overall, we use Panoptic Quality (PQ)~\cite{8953237}, which combines detection (RQ - Recognition Quality) and segmentation (SQ - Segmentation Quality) performance.

\subsection{Results}
\label{sec:results}

\paragraph{Quantitative.}  
Because of the limited number of training data, our algorithm achieves \textbf{78.4}\% AP@0.25, \textbf{63.7}\% AP@0.50, \textbf{67.2}\% mIoU, and \textbf{64.5}\% PQ on the real-world testing dataset.  

\begin{figure}[ht]
    \centering
    {\includegraphics[width=0.25\textwidth]{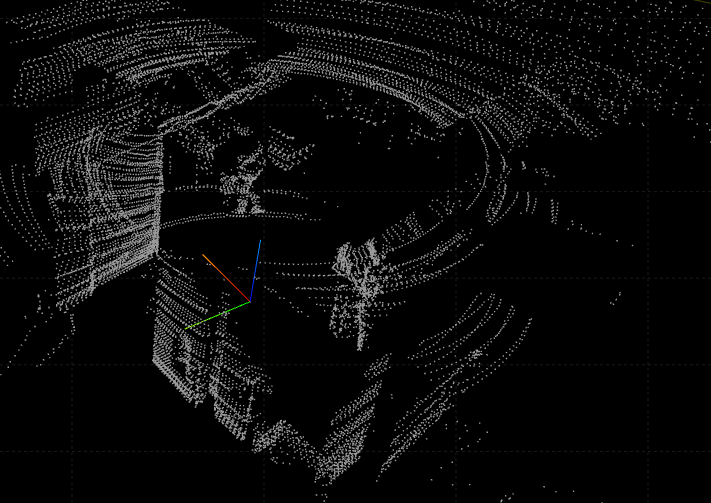}\label{subfig:image1}}
    \hfill
    {\includegraphics[width=0.2\textwidth]{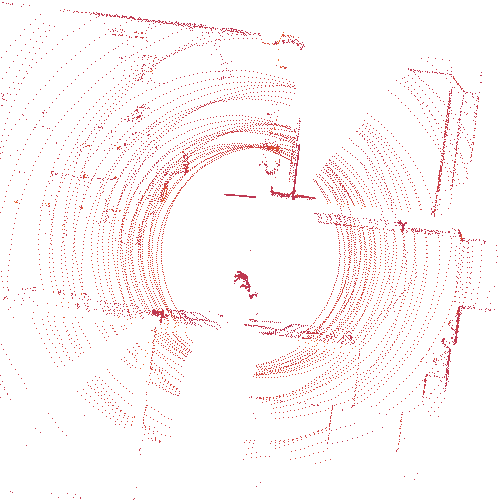}\label{subfig:image2}}
    \hfill
    {\includegraphics[width=0.2\textwidth]{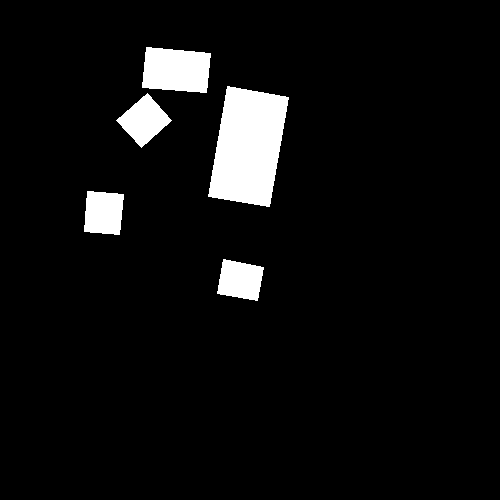}\label{subfig:image3}}
    \hfill
    {\includegraphics[width=0.2\textwidth]{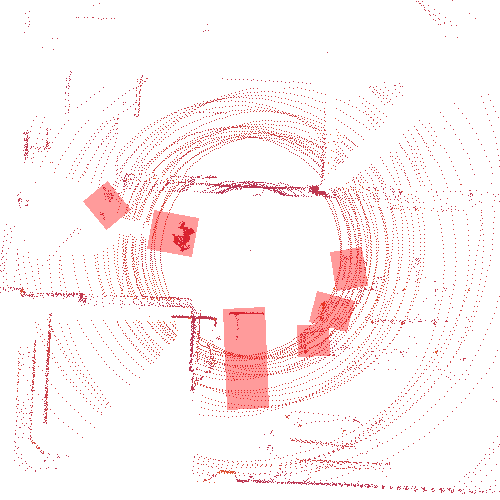}\label{subfig:image4}}
    \hfill
    \\
    {\includegraphics[width=0.25\textwidth]{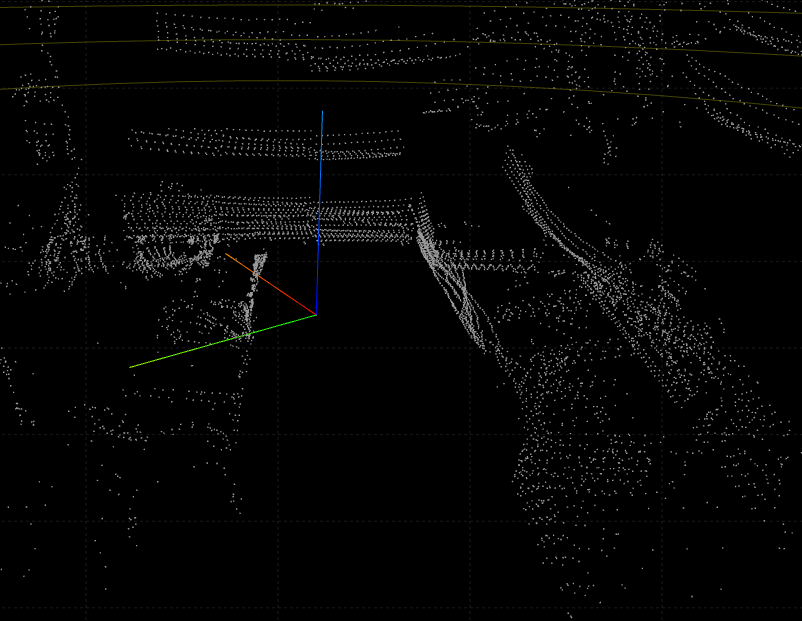}\label{subfig:image5}}
    \hfill
    {\includegraphics[width=0.2\textwidth]{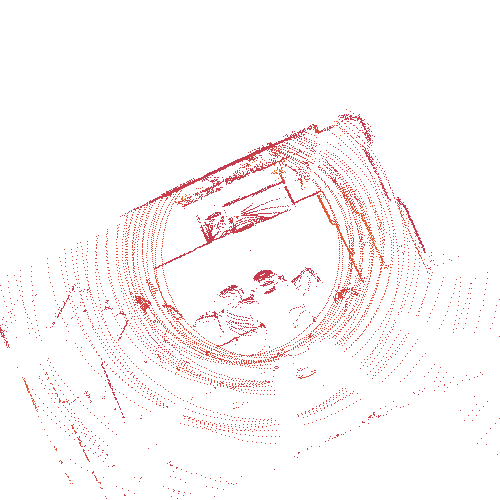}\label{subfig:image6}}
    \hfill
    {\includegraphics[width=0.2\textwidth]{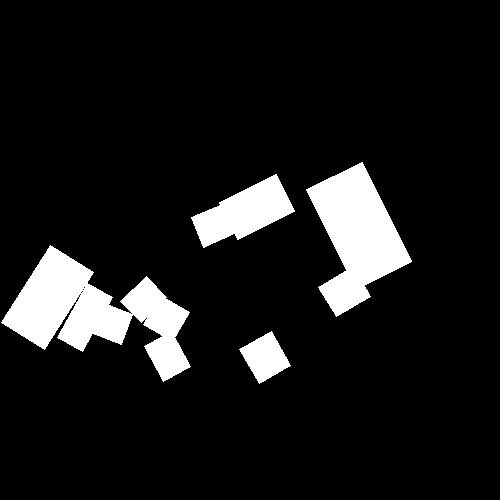}\label{subfig:image7}}
    \hfill
    {\includegraphics[width=0.2\textwidth]{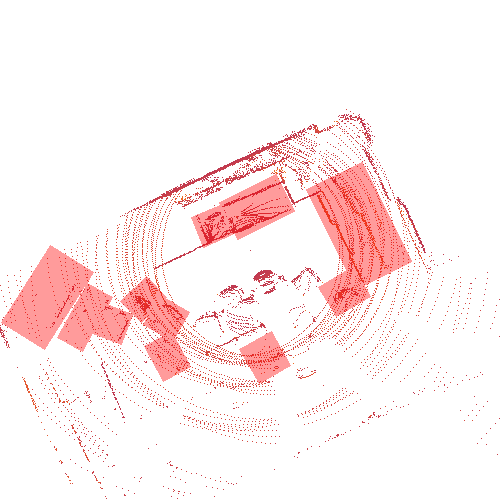}\label{subfig:image8}}
    \hfill
    \\
    {\includegraphics[width=0.25\textwidth]{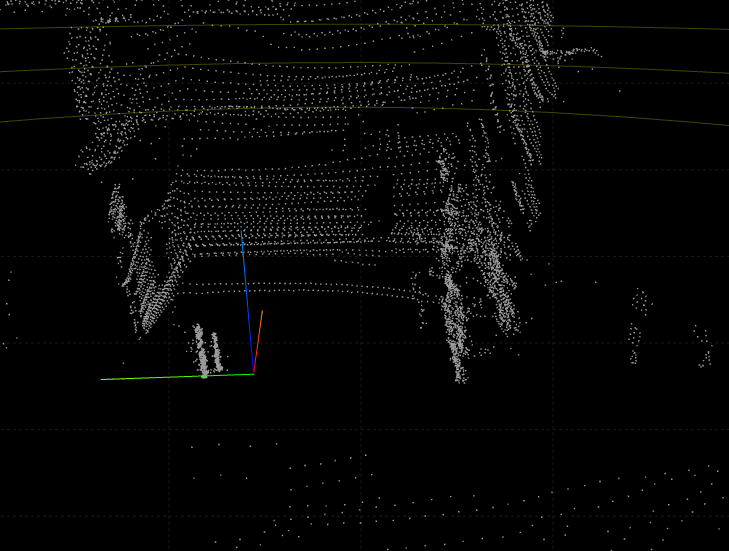}\label{subfig:image9}}
    \hfill
    {\includegraphics[width=0.2\textwidth]{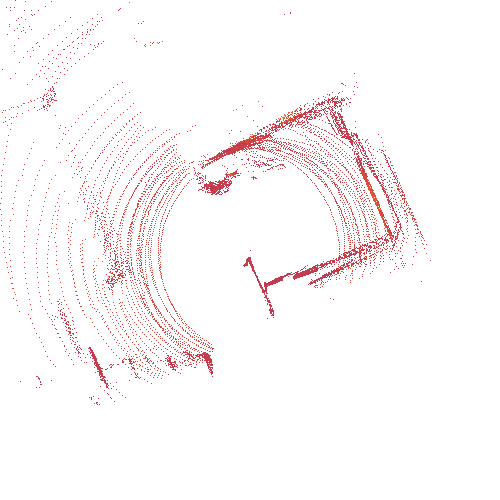}\label{subfig:image10}}
    \hfill
    {\includegraphics[width=0.2\textwidth]{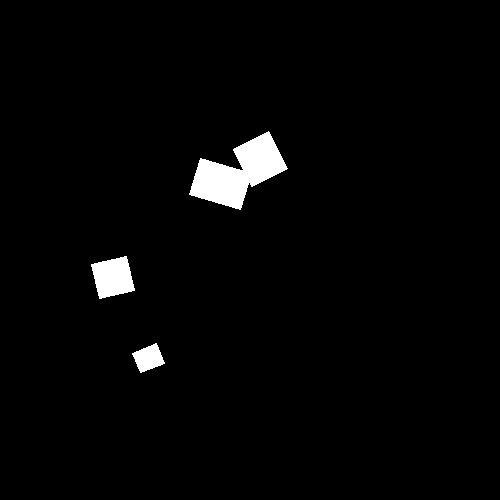}\label{subfig:image11}}
    \hfill
    {\includegraphics[width=0.2\textwidth]{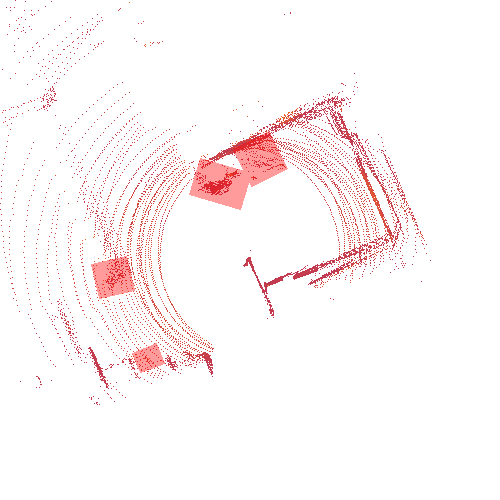}\label{subfig:image12}}
    \\
    {\includegraphics[width=0.25\textwidth]{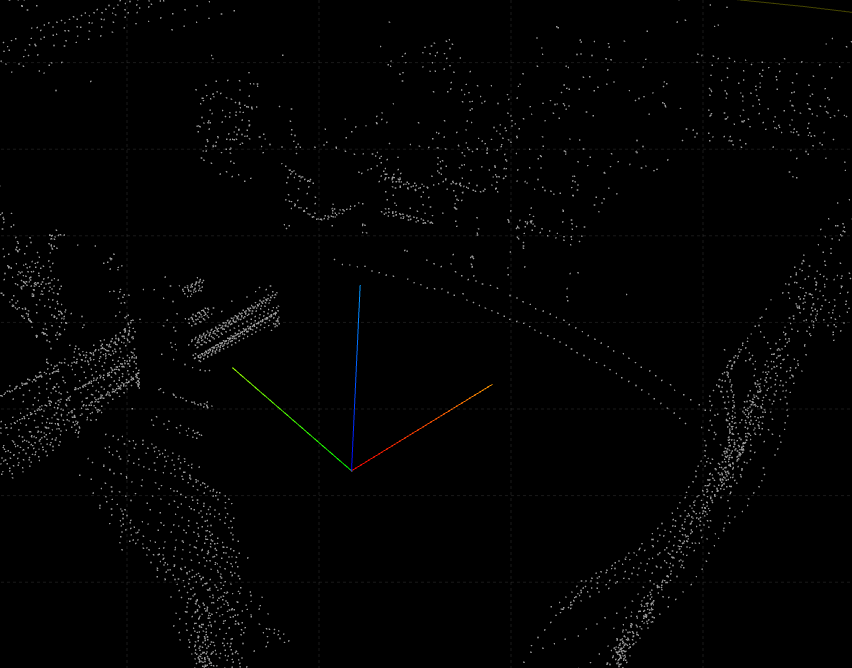}\label{subfig:image9}}
    \hfill
    {\includegraphics[width=0.2\textwidth]{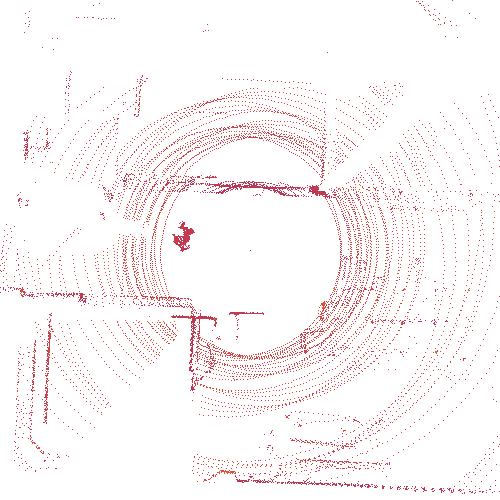}\label{subfig:image10}}
    \hfill
    {\includegraphics[width=0.2\textwidth]{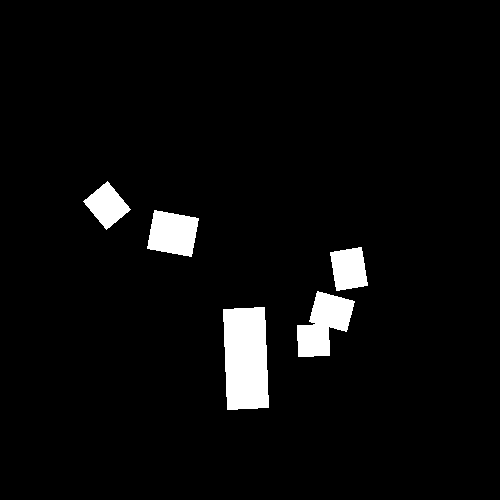}\label{subfig:image11}}
    \hfill
    {\includegraphics[width=0.2\textwidth]{figs/r4.png}\label{subfig:image12}}
    \\
    {\includegraphics[width=0.25\textwidth]{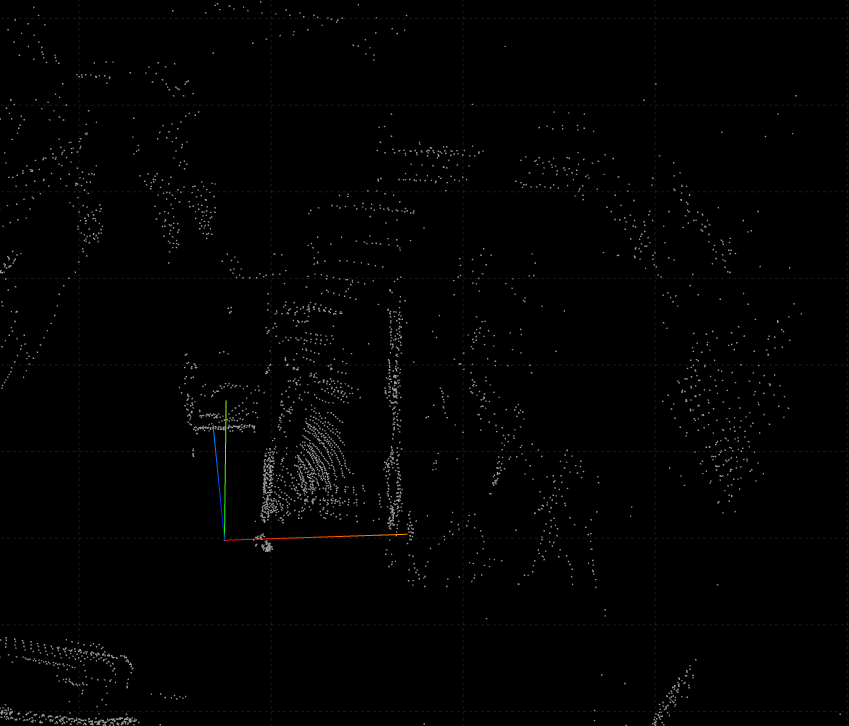}\label{subfig:image9}}
    \hfill
    {\includegraphics[width=0.2\textwidth]{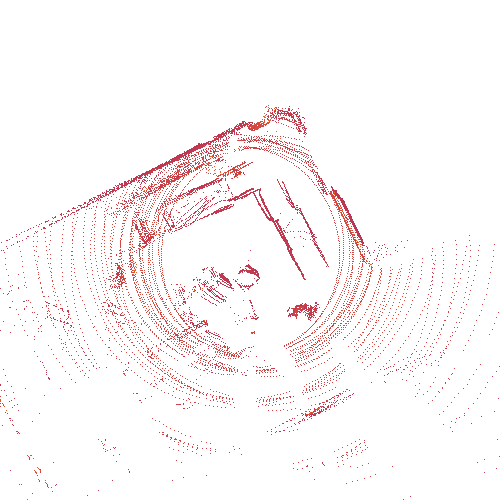}\label{subfig:image10}}
    \hfill
    {\includegraphics[width=0.2\textwidth]{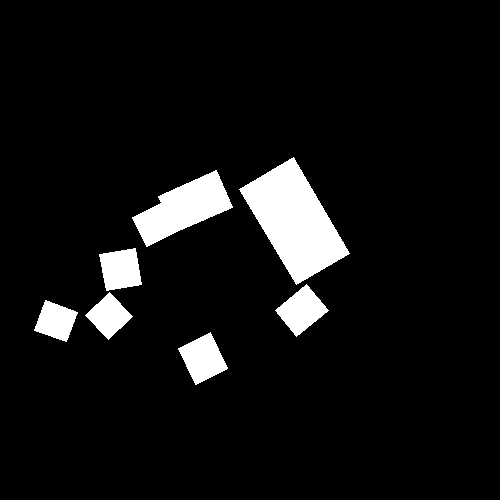}\label{subfig:image11}}
    \hfill
    {\includegraphics[width=0.2\textwidth]{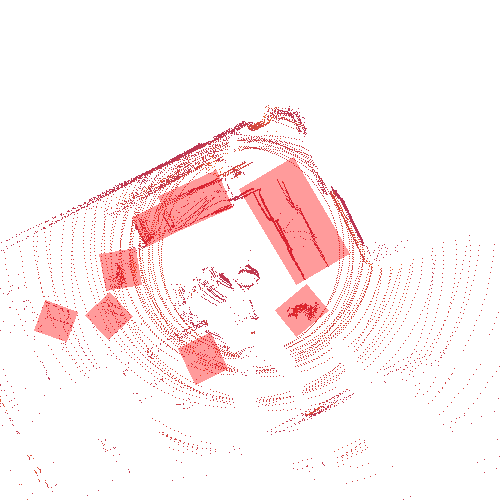}\label{subfig:image12}}
    \\
    {\includegraphics[width=0.25\textwidth]{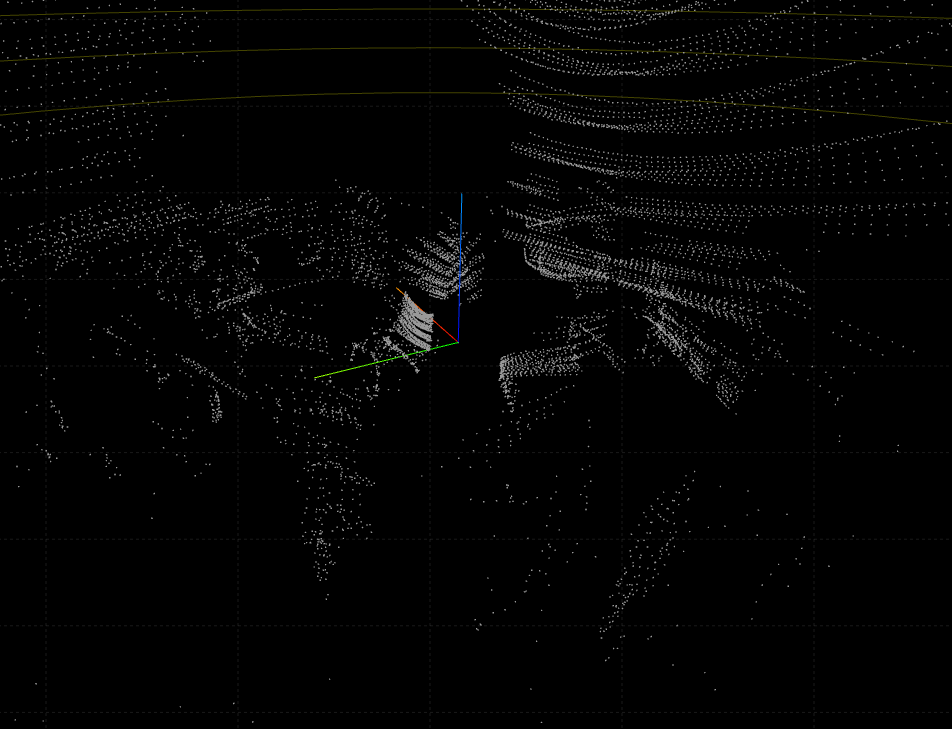}\label{subfig:image9}}
    \hfill
    {\includegraphics[width=0.2\textwidth]{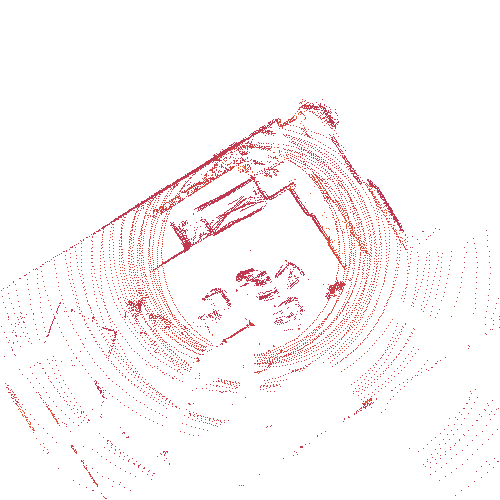}\label{subfig:image10}}
    \hfill
    {\includegraphics[width=0.2\textwidth]{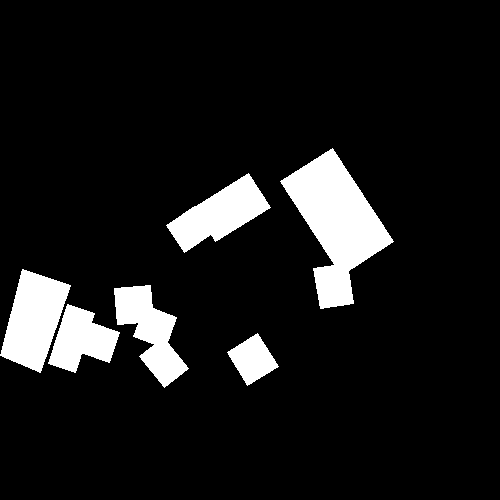}\label{subfig:image11}}
    \hfill
    {\includegraphics[width=0.2\textwidth]{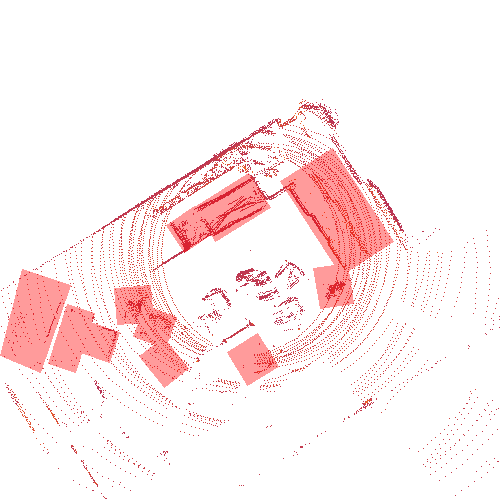}\label{subfig:image12}}
    \caption{Results of IndoorBEV: The first column represents the raw 3D point clouds as each input. The second column is the axis fusion result of each correlated input. The third column illustrates the predicted mask of each objects in each secnes. The last column is the final BEV result of each input}
    \label{fig:resultsmore}
\end{figure}

\paragraph{Qualitative.}  
Fig.~\ref{fig:teaser} and Fig.~\ref{fig:resultsmore} show qualitative results on representative scenes from the test set. The visualizations include the input point cloud BEV projection, the ground truth masks, and the predicted masks overlaid on the BEV. These results demonstrate the model's ability to accurately detect diverse objects (both static furniture and dynamic elements like robots) and precisely delineate their footprints, even in cluttered scenes with occlusions. The mask predictions capture object shapes more faithfully than traditional bounding boxes.

\section{Conclusion}
\label{sec:conclusion}

We presented a novel IndoorBEV framework designed for robust perception in complex indoor environments using lidar point clouds. Our approach addresses the limitations of traditional 3D object detection methods, particularly their reliance on bounding boxes which struggle with the diverse shapes and clutter typical of indoor settings.
Our query-based decoder head, which directly predicts instance footprint masks in the BEV space alongside object classes and 3D pose attributes. This mask-centric formulation bypasses the need for bounding box regression, allowing the framework to accurately capture the 2D spatial extent of varied objects without strong geometric priors. By predicting masks, IndoorBEV provides a representation that is directly beneficial for downstream robotic tasks such as collision checking, path planning, and interaction.
Our experiments, conducted on a hybrid dataset combining simulated and real-world indoor lidar scans, demonstrate the efficacy of IndoorBEV. Furthermore, IndoorBEV allows to operate in real-time, utilizing its suitability for deployment on mobile robot platforms for on-the-fly perception and decision-making is possible.
This work pioneers the application of  BEV perception for comprehensive indoor scene understanding on mobile robots. It offers a unified solution for detection and segmentation, providing a valuable perception module for autonomous systems navigating challenging indoor spaces. Future research directions include incorporating temporal information for enhanced tracking of dynamic objects, exploring multi-modal fusion with visual data within the BEV framework, and further optimization for deployment on computationally constrained hardware.

\acknowledgments{This research work has received funding from the European Commission’s HORIZON Marie Skłodowska-Curie Actions (MSCA) under Grant agreement No. 101072634, project RAICAM.}




\section{Limitations}
\label{sec:limitations}

Although IndoorBEV achieves promising results, several limitations persist that highlight important directions for future improvement:

\begin{itemize}
    \item \textbf{Sparse Point Cloud Inputs.}  
    The MID-360 LiDAR sensor adopted for our mobile robot platform offers high portability and a wide field of view but produces relatively low-density point clouds. This sparsity reduces the resolution of environmental representation and can negatively impact detection and segmentation performance, especially in complex or cluttered indoor spaces. Future work could address this limitation by integrating higher-resolution sensors or employing multi-sensor fusion techniques to achieve denser and more informative scans without sacrificing mobility.

    \item \textbf{Limited Real-World Dataset Diversity.}  
    Our real-world training data was collected from only five indoor environments, covering a restricted range of spatial layouts, object categories, and environmental conditions. This relatively small and homogeneous dataset may limit the generalization capability of the model when deployed in unseen or highly varied indoor scenes. Expanding the dataset to include more diverse settings, dynamic objects, and environmental variations is crucial for improving robustness.

    \item \textbf{Computational Constraints on Mobile Platforms.}  
    Mobile robots typically operate under tight computational and power budgets, limiting the feasible complexity of onboard perception models. Although IndoorBEV is designed for real-time processing, trade-offs between model size, speed, and accuracy may arise, potentially leading to delayed responses or reduced precision during navigation. To mitigate this, future work should explore model compression, quantization, and hardware-accelerated inference techniques to maintain high performance within resource-constrained environments.
\end{itemize}

By systematically addressing these limitations, IndoorBEV can be further developed for deployment in diverse, dynamic, and computationally constrained indoor robotic applications.

\section{Future Work}
\label{sec:future}

Building upon the foundation established by IndoorBEV, several directions are planned for future research:

\begin{itemize}
    \item \textbf{Expanded Real-World Data Collection.}  
    We will extend our dataset to include a broader range of indoor layouts, object categories, and environmental conditions, using various mobile platforms equipped with different sensor configurations.

    \item \textbf{Broader Comparative Evaluation.}  
    Future work will benchmark IndoorBEV against a wider range of baselines, including classical LiDAR-based detectors, BEV segmentation frameworks, and real-time-oriented indoor perception systems.

    \item \textbf{Real-World Deployment and Optimization.}  
    We aim to integrate IndoorBEV into a complete real-time robotic navigation stack, transitioning from simulation to physical deployment. This will involve optimizing onboard inference through techniques such as model pruning, quantization, and pipeline acceleration to ensure low-latency, closed-loop operation.
\end{itemize}

These future efforts aim to further advance BEV-based indoor perception toward robust, real-world deployment on autonomous mobile robots.

\section{Appendix}

\subsection{Data Augmentation}  
Extensive preprocessing and augmentation are applied:
\begin{itemize}
    \item \textbf{Point Cloud Augmentation:} Random shuffling, rotation, decimation, jittering, global noise injection, and random point dropping.
    \item \textbf{Object-Level Augmentation:} Box-level noise on location, size, and yaw; object sampling and insertion from a prebuilt object database with collision checking.
    \item \textbf{Mask Augmentation:} Additional perturbations on BEV masks to simulate label noise.
\end{itemize}

\subsection{Analytic forms for ray–object intersection tests}
\label{app2}
\begin{table}[h]
\centering
\renewcommand{\arraystretch}{1.2}
\setlength{\tabcolsep}{8pt}
\begin{tabularx}{\textwidth}{c X}
\toprule
\textbf{Geometry} & \textbf{Intersection Equation} \\
\midrule
Plane & $\mathbf{n} \cdot (\mathbf{o} + t \mathbf{d}) + d = 0$ \\
Sphere & Solve $a t^2 + b t + c = 0$, where $a = \|\mathbf{d}\|^2$, $b = 2 \mathbf{o} \cdot \mathbf{d}$, $c = \|\mathbf{o}\|^2 - r^2$ \\
Box & Slab method: intersection intervals along each axis \\
Cylinder & Quadratic equation in $x$–$y$ plane, plus height bounds in $z$ \\
Ellipsoid & Transform to unit sphere space, then solve as a sphere \\
Capsule & Combination of cylinder and hemisphere intersections \\
\bottomrule
\end{tabularx}
\caption{Analytical expressions for ray–object intersection tests.}
\label{tab:intersection}
\end{table}

\subsection{IndoorBEV: Cluster Encoding and Query-Based Mask Prediction Algorithm}
\label{app1}
\begin{algorithm}[h]
\small
\caption{IndoorBEV: Cluster Encoding and Query-Based Mask Prediction}
\label{alg:IndoorBEV}
\textbf{Input}: Batch of Point Clouds $\{P_b\}_{b=1}^B$, where $P_b = \{p_i\}_{i=1}^{N_b}$, $p_i \in \mathbb{R}^d$. \\
\textbf{Parameters}: Cluster size $v$, spatial ranges $(x_{range}, y_{range}, z_{range})$, number of queries $N_q$. \\
\textbf{Modules}: Cluster-division $\mathcal{V}$, AF-PCCN $\mathcal{E}_{pillar}$, BEVScatter $\mathcal{S}_{bev}$, Backbone $\mathcal{B}$, Upsampler $\mathcal{U}$, DecoderHead $\mathcal{H}_{dec}$. \\
\textbf{Output}: Detections per batch item: $\{ \text{logits}_j, \text{dims}_j, \text{pos}_j, \text{yaw}_j, \text{mask}_j \}_{j=1}^{N_q}$. \\
\textbf{BEGIN}:
\begin{algorithmic}[1]
\STATE Initialize empty lists for batched clusters $V_{batch}$, coordinates $C_{batch}$, num\_points $N_{p\_batch}$.
\FOR{each point cloud $P_b$ in batch}
    \STATE $P_b^{filtered} \leftarrow \text{FilterInRange}(P_b, x_{range}, y_{range}, z_{range})$ \hfill \textit{\% Remove points outside range}
    \STATE $V_b, C_b, N_{p\_b} \leftarrow \mathcal{V}(P_b^{filtered})$ \hfill \textit{\% Clustered points}
    \STATE Append $V_b$ to $V_{batch}$, $N_{p\_b}$ to $N_{p\_batch}$.
    \STATE $C_{b\_padded} \leftarrow \text{PadBatchIndex}(C_b, b)$ \hfill \textit{\% Add batch index to coordinates}
    \STATE Append $C_{b\_padded}$ to $C_{batch}$.
\ENDFOR
\STATE $V \leftarrow \text{Concatenate}(V_{batch})$ \hfill \textit{\% Combine clusters from batch}
\STATE $C \leftarrow \text{Concatenate}(C_{batch})$ \hfill \textit{\% Combine coordinates}
\STATE $N_p \leftarrow \text{Concatenate}(N_{p\_batch})$ \hfill \textit{\% Combine point counts}

\STATE $F_{axis} \leftarrow \mathcal{E}_{compact}(V, N_p, C)$ \hfill \textit{\% Extract features per compact content}
\STATE $F_{bev} \leftarrow \mathcal{S}_{bev}(F_{axis}, C, B)$ \hfill \textit{\% Scatter axis features to BEV grid}
\STATE $F_{bev} \leftarrow \text{LayerNorm}(F_{bev})$ \hfill \textit{\% Normalize BEV map}

\STATE $F_{backbone} \leftarrow \mathcal{B}(F_{bev})$ \hfill \textit{\% Process BEV map with backbone}
\STATE $F_{upsampled} \leftarrow \mathcal{U}(F_{backbone})$ \hfill \textit{\% Upsample final backbone features}

\STATE $L_{cls}, D, P, Y, M \leftarrow \mathcal{H}_{dec}(F_{upsampled})$ \hfill \textit{\% Pass features through query-based decoder head}
    \STATE \textit{\% Inside $\mathcal{H}_{dec}$:}
    \STATE \textit{\% \quad Project $F_{upsampled}$ to $F_{proj}$}
    \STATE \textit{\% \quad Initialize learnable queries $Q$}
    \STATE \textit{\% \quad Refine queries: $Q' \leftarrow \text{SimpleHeadDecoder}(Q, F_{proj})$}
    \STATE \textit{\% \quad Predict logits, dims, pos, yaw from $Q'$}
    \STATE \textit{\% \quad Generate mask embeddings $M_{embed}$ from $Q'$}
    \STATE \textit{\% \quad Compute mask logits $M = M_{embed} \cdot F_{proj}$}

\RETURN $\{L_{cls}, D, P, Y, M\}$ \hfill \textit{\% Return dictionary of predictions}
\end{algorithmic}
\textbf{END}
\end{algorithm}

\bibliography{example}  

\end{document}